# Identifying the number of clusters for K-Means: A hypersphere density based approach

Sukavanan Nanjundan, Shreeviknesh Sankaran, C.R. Arjun, G. Paavai Anand
Department of Computer Science and Engineering
SRM Institute of Science and Techonology
Chennai, India
n.sukavanan@gmail.com, shreeviknesh@gmail.com,
nujrachembai@gmail.com, pavai_gops@yahoo.co.in

*Abstract*—Application of the K-Means algorithm is restricted by the fact that the number of clusters should be known beforehand. Previously suggested methods to solve this problem are either ad hoc or require parametric assumptions and complicated calculations. The proposed method aims to solve this conundrum by considering cluster hypersphere density as the factor to determine the number of clusters in the given dataset. The density is calculated by assuming a hypersphere around the cluster centroid for n-different number of clusters. The calculated values are plotted against their corresponding number of clusters, and then the optimum number of clusters is obtained after assaying the elbow region of the graph. The method is simple and easy to comprehend and provides robust and reliable results.

## I. INTRODUCTION

K-Means is an algorithm used for clustering, and it is popular for performing cluster analysis in data mining [1]. K-Means clustering is used to divide or distribute n observations into k clusters in which each observation belongs to the cluster with the nearest centroid, which serves as a prototype of the cluster [2][3]. The number of clusters is one of the inputs required for this algorithm, which is hard to determine beforehand since K-Means is generally used for unsupervised learning.

The optimal number of clusters is a prerequisite because if the number of clusters given as input to the K-Means algorithm is fewer than the optimal value. In such a case, the algorithm will produce a result that does not capture the important aspects or the essence of the underlying data. In contrast, if the assumed K value is greater than the optimal value, then the model built will represent unnecessary associations between data points.

This issue is traditionally overcome by trial-and-error, wherein the cluster purity [4] is measured for many values of K, which is both computationally inefficient and time-consuming.

There are several methods available to identify the optimal number of clusters for a given dataset, but only a few provide reliable and accurate results such as the Elbow method [5], Average Silhouette method [6], Gap Statistic method [7]. Sometimes even these methods provide different results for the same dataset. In addition, some of these methods involve complex computations and require a lot of time to provide results.

Our approach towards this problem is simplistic and provides a reliable solution. The optimum number of clusters is determined by plotting a graph of cluster hypersphere density vs. the number of clusters for different values of the number of clusters.

Given a well-distributed data set, it is observed that the mean cluster hypersphere density of the clusters decreases with an increase in the number of clusters in a non-linear fashion. The resulting graph looks similar to that of the graph obtained in Elbow method wherein the decrease in mean cluster hypersphere density is rapid when K is less than the optimal value and gradually decreases as it nears the optimum number of clusters, after which the gradient becomes almost constant or the graph changes direction. This region is known as the "Elbow" region. Amongst the points in the elbow region lies the optimum number of clusters. Sometimes, the elbow region contains a high range of values. In this scenario, coupling this algorithm with pre-existing methods such as the Average Silhouette method or any of the available methods will provide the required output.

Implementation of our method results in identifying the optimal number of clusters for the given dataset. The result is both precise and reliable, and sometimes produces better results than some of the existing methods, which is proved by the figures given below. Finding the optimal number of clusters is extremely crucial in order to represent the underlying relationship between datapoints and attributes perfectly, and our method helps in finding optimal values.

## II. SOME OF THE EXISTING APPROACHES

There are some methods or approaches to deal with the problem of finding the optimal number of clusters for a given dataset, of which the most prominent methods have been discussed here.

### A. Average Silhouette Method

In the average silhouette method, a silhouette value for every datapoint is calculated, the mean of which is used to find the optimal number of clusters. The silhouette value represents how similar a datapoint is to its own cluster when compared to all the other clusters or cluster centroids. The value ranges from -1 to +1. A higher silhouette value implies that the datapoint is matched well to its own centroid/cluster and is not so well matched with other clusters. If the mean of the silhouette value measured for all the datapoints is considerably high, then it can be said that the number of clusters is at its optimal value, or in other words, the clustering structure is appropriate. On the other hand, if the mean silhouette value turns out to be very less or negative, then it means that the cluster structure is not proper, and it may be having either more or lesser number of clusters than the optimal value. To find the silhouette value, any distance metric, like Minkowski distance or Euclidean distance, can be used.

$$s(i) = \frac{b(i) - a(i)}{max\{a(i), b(i)\}}$$

*B. Elbow Method*

In the elbow method, the variance (within-cluster sum of squared errors) is plotted against the number of clusters. The first few clusters will introduce a lot of variance and information, but at some point, the information gain will become low, thus imparting an angular structure to the graph. The optimal number of clusters is found out from this point; therefore, this is known as the "elbow criterion." But this point cannot always be determined without any sense of ambiguity.

*C. Gap Statistics*

The gap value (gap statistic) is the difference between the within-cluster dispersion for different values of k and their expected values. The value of k that maximizes the gap statistic (i.e., the value that yields the minimum intra-cluster variation under an appropriate null distribution) is the estimated optimal number of clusters.

$$Gap(k) = \frac{1}{B}\sum_{b=1}^{B} log(W_{kb}^*) - log(W_k)$$

Figure 1 – Gap Statistics Formula

where B is the number of reference datasets with random uniform distribution, k is the number of clusters, $W_k$ is the total intra-cluster variation for the observed datasets, and $W_{kb}$ is the total intra-cluster variation for the reference datasets.

III. ALGORITHM

*A. K-Means Algorithm*

The K-means clustering algorithm, as mentioned earlier, is used to distribute n datapoints over k different clusters. The number of clusters or 'k' value and the dataset is given as input to the algorithm. The dataset is a collection of values for a particular set of attributes for each data point. The algorithm starts by selecting some points to represent the 'k' centroids, which can either be randomly generated or even manually selected from the dataset. The algorithm then iterates between two steps.
- Assign a datapoint to the nearest centroid.

Recalculate the mean centroid after assigning datapoint. The distance measure used in our approach is Euclidean, whereas other distance measures like Hamming distance, Chebyshev distance, etc., can also be used.

```
Randomly initialize K cluster centroids:
Repeat {
    for i = 1 to m
        c⁽ⁱ⁾ := index (from 1 to K) of cluster centroid
        closest to x⁽ⁱ⁾
    for k = 1 to K
        u_k := average (mean) of points assigned to
        cluster k
}
note: m is number of samples
```

Figure 2 – K-Means Algorithm

*B. Cluster Hypersphere Density-Based Algorithm*
- Run K-Means algorithm on the data set 'n' times, where the number of clusters varies from 1 to n (n is a randomly chosen number, which is chosen intuitively).
- For each value of K, note the metrics produced by the silhouette score method and elbow method.
- Calculate individual cluster densities by considering the clusters to be hyperspheres.

  o $Volume = \frac{\pi^{\frac{n}{2}}}{\Gamma(\frac{n}{2}+1)} R^n$
  o where $\Gamma$ is Leonhard Euler's gamma function, $\Gamma(n) = (n-1)!$, n is the number of dimensions, and R is the radius of the hypersphere.
  o Radius for the cluster is calculated as the distance between the cluster centroid and the farthest point from it, which belongs to that particular cluster.
  o $Density = \frac{Volume\ of\ the\ cluster}{Number\ of\ points\ in\ the\ cluster}$

- Plot the mean density of clusters against the corresponding number of clusters.
- From the resulting graph, the optimum number of cluster values can be discerned from the elbow region.

IV. RESULTS

All the algorithms are implemented in python 3.7.0, and the experiments were carried out on an i5 8th gen 1.8 GHz machine with 8 GB RAM and WINDOWS-10 platform.

*A. Iris Dataset*

The Iris dataset contains information pertaining to 3 different types of irises (Setosa, Versicolor, and Virginica) such as petal length, sepal length, petal width, and sepal width stored in a matrix. The columns represent the 4 attributes mentioned above, and the rows represent the values of those 4 attributes for 150 datapoints.

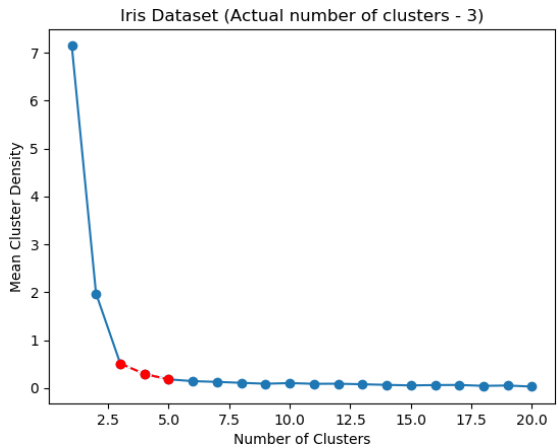

*Figure 3 – Hypersphere density graph for Iris Dataset*

Clearly (from Fig. 3,) the elbow region resides in the range of 3 – 5 (number of clusters.) Since the cluster hypersphere density does not decrease significantly beyond 3, we can conclude that 3 is the optimal number of clusters.

*B. Synthetic Dataset – I*

This is a synthesized 2-d dataset with 3000 data points, which are equally distributed over 20 cluster centroids.

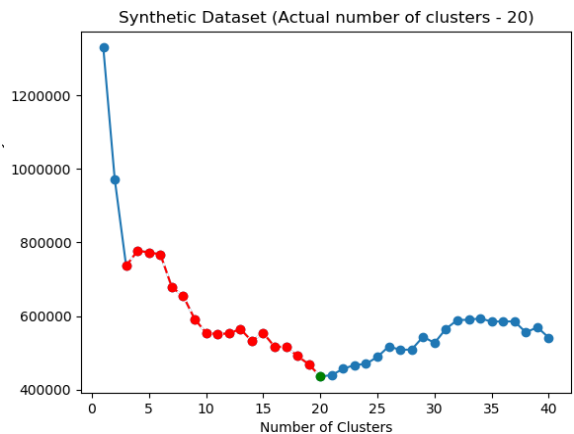

*Figure 4 – Hypersphere density graph for Synthetic 2-d Dataset*

This is a peculiar case in which the elbow extends for a long-range (between 3 and 20). But, 20 can be chosen as the optimal value as it is apparent that the cluster hypersphere density reaches a global minimum value, and there is an apparent change in the direction of the graph.

*C. Breast Cancer Dataset*

This is a normalized version of the breast cancer dataset consists of 2 classes of cancer, namely, malignant and benign cancer. There are a total of 569 datapoints where each datapoint is defined by 30 attributes.

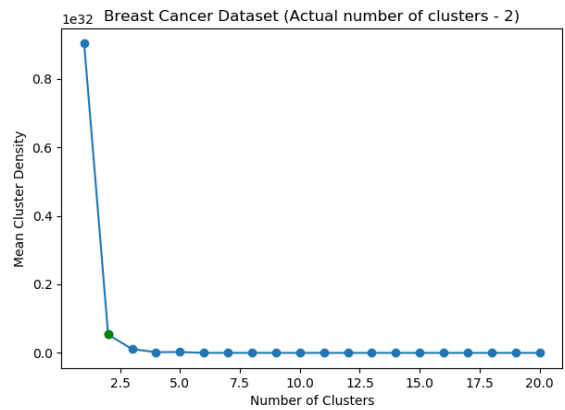

*Figure 5 - Hypersphere density graph for Breast Cancer Dataset*

There is a conspicuous change in cluster hypersphere density value from K=1 to K=2, after which the change in cluster hypersphere density becomes negligible. This implies that 2 is the optimal number of clusters.

*D. Cars Dataset*

This dataset consists of information about 3 brands/make of cars, namely the US, Japan, Europe. There are 261 datapoints, each of which is defined by 8 attributes.

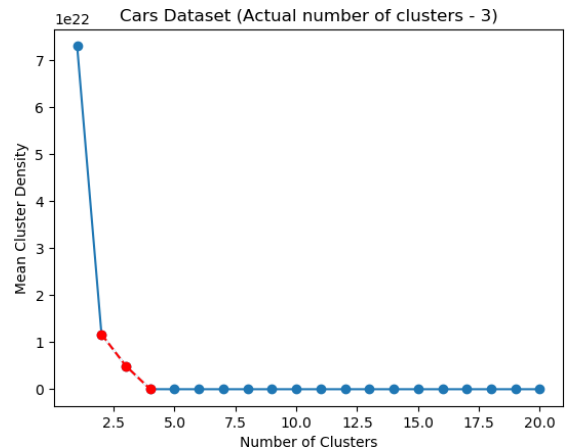

*Figure 6 - Hypersphere density graph for Cars Dataset*

The elbow region for this dataset exists between K=2 and K=4, as highlighted in the graph. But the change in cluster hypersphere density from K=2 to K=3 and from K=3 to K=4 is nearly the same. Therefore it is hard to pinpoint a value as the optimal number of clusters. Thus, in this case, any other method to determine the number of clusters (such as average silhouette and elbow methods) can be combined with our method to find out the optimal number of clusters.

*E. Synthetic Dataset – II*

This is a synthesized 6-d (6 attributes) dataset wherein 5000 datapoints have been distributed equally over 10 cluster centroids.

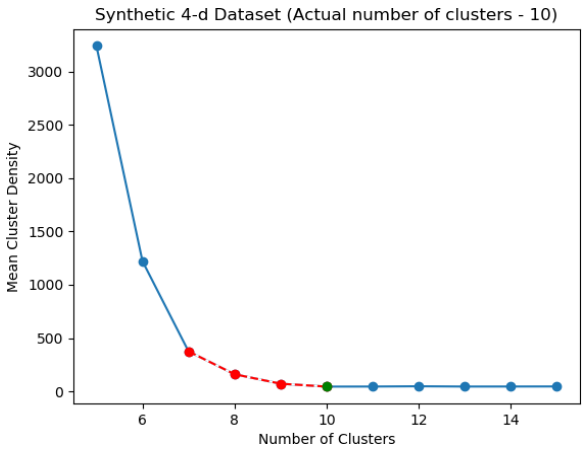

*Figure 7 - Hypersphere density graph for Synthetic 4-d Dataset*

The elbow region in this graph exists between K=6 and K=10, beyond which the change in cluster hypersphere density can be considered as negligible. In addition to that, the mean cluster hypersphere density is minimum for K=10, which is the optimal number of clusters. The graph has been zoomed in to represent the elbow region clearly.

### F. IPL Dataset

In this dataset, there are 87 datapoints which represent the normalized version of batting and bowling averages of IPL players.

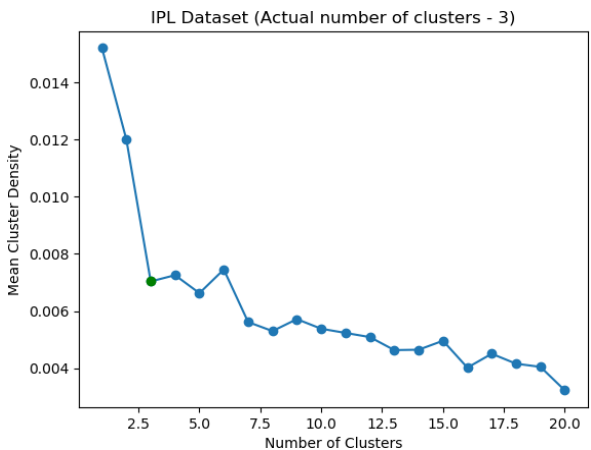

*Figure 8 - Hypersphere density graph for IPL Dataset*

This is another peculiar case where there is no elbow region as such, but there exists an elbow point (K=3, as highlighted in the graph), which corresponds to the optimal number of clusters. The mean cluster hypersphere density drop is very high between points K=1 to K=3, beyond which the drop in mean cluster hypersphere density is very less when compared to the drops before the point K=3. Even though the number of datapoints is very less, our method is able to determine the optimal number of clusters correctly.

## V. COMPARISONS

### A. Iris Dataset

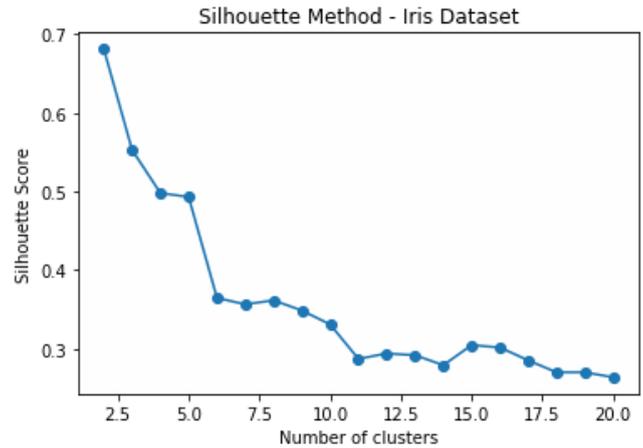

*Figure 9 – Silhouette Value Graph for Iris Dataset*

From figure 9, it is clear that the high value of silhouette score at K=2 indicates that the optimum number of clusters should 2, whereas, in reality, that is not the case. While the silhouette score fails to identify the optimal number of clusters, our method accurately predicts the optimal value, as shown in figure 3.

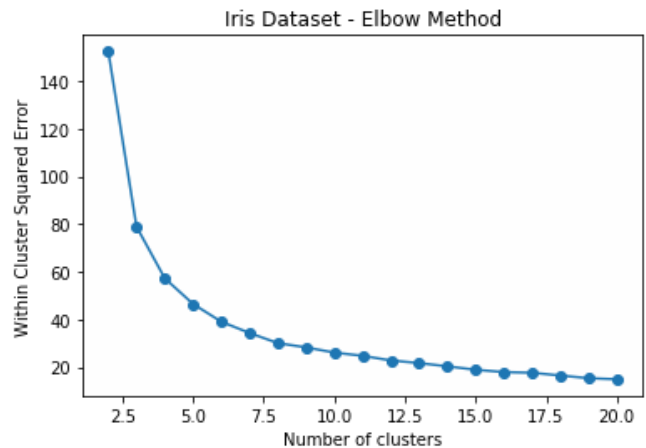

*Figure 10 – Elbow graph for Iris Dataset*

While the silhouette method is not effective in identifying the number of clusters for the iris dataset, the elbow method is better in comparison. The drop in squared error is high from K=2 to K=3, whereas after that, the drop tends to be smaller. Thus it can be inferred that the optimal number of clusters is 3.

### B. Synthetic Dataset - I

In figure 11, the silhouette score is maximum at K=20, thus implying that 20 is the optimal number of clusters. This implication is in line with the real value of the number of clusters, which is 20.

From the output of the elbow method (figure 12), it is apparent that the elbow region extends over a long range of values, just like it did for our method (figure 4). But the result in our method showed that the mean cluster hypersphere density was minimum at K=20, which is the optimal value. There is no such point in the elbow method and even after zooming in, which is represented in figure 13, the elbow point seems to be occurring at K=19, which is not the right value.

Thus, in this case, the silhouette score method is able to identify the optimal number of clusters, whereas the elbow method fails to do so.

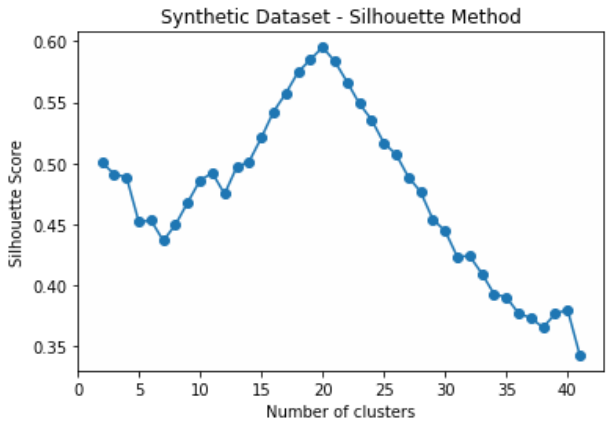

*Figure 11 – Silhouette Value Graph for Synthetic 2-d Dataset*

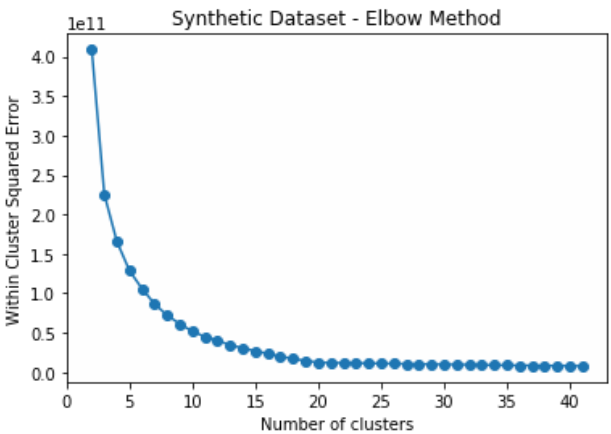

*Figure 12 – Elbow Graph for Synthetic 2-d Dataset*

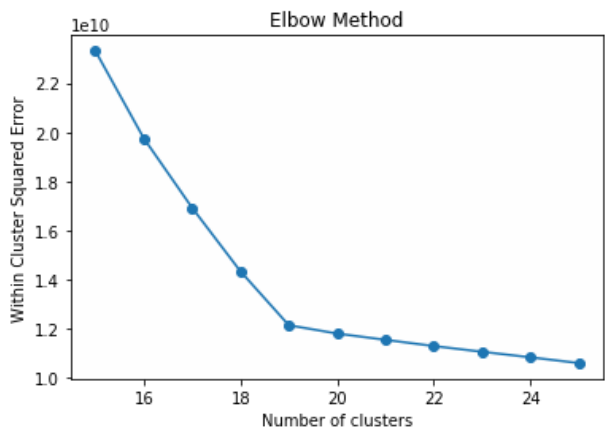

*Figure 13 – Elbow Graph for Synthetic 2-d Dataset (Zoomed In)*

## VI. CONCLUSION

This research introduces a new method, which is both simple and effective, to identify the optimal number of clusters for the K-Means algorithm. In our method, the clusters are assumed to be in the shape of hyperspheres, whose mean densities are used as the factor to determine the optimal number of clusters.

While assuming the shape of the clusters to be hyperspheres may seem wrong, our method proves that it works very well by accurately finding the number of clusters for the Iris dataset, where there is a huge overlap between 2 of the 3 clusters present. Even the silhouette method fails to identify the optimal number of clusters, but our method holds true.

In the case of a synthetic dataset, our method is much more effective where it produces clear optimal value wherein the mean cluster hypersphere density is minimum at that point. As shown in figures 11, 12, and 13, the silhouette score method is successful in identifying the number of clusters, whereas the elbow method fails.

Thus, from the observations presented above, it can be concluded that our method provides an elbow region for real-time datasets (which are not perfectly distributed) from which the optimal number of clusters can be identified by analyzing the drops in mean cluster hypersphere density from one K value to the other. In the case of synthetic data or well-distributed data, the mean cluster hypersphere density reaches a minimum value for a particular K value, and this K value is the optimal number of clusters.